# An Idiotypic Immune Network as a Short-Term Learning Architecture for Mobile Robots

Amanda Whitbrook, Uwe Aickelin, and Jonathan Garibaldi

School of Computer Science, University of Nottingham, UK, NG8 1BB
{amw,uxa,jmg}@cs.nott.ac.uk

**Abstract.** A combined Short-Term Learning (STL) and Long-Term Learning (LTL) approach to solving mobile robot navigation problems is presented and tested in both real and simulated environments. The LTL consists of rapid simulations that use a Genetic Algorithm to derive diverse sets of behaviours. These sets are then transferred to an idiotypic Artificial Immune System (AIS), which forms the STL phase, and the system is said to be seeded. The combined LTL-STL approach is compared with using STL only, and with using a hand-designed controller. In addition, the STL phase is tested when the idiotypic mechanism is turned off. The results provide substantial evidence that the best option is the seeded idiotypic system, i.e. the architecture that merges LTL with an idiotypic AIS for the STL. They also show that structurally different environments can be used for the two phases without compromising transferability.

## 1 Introduction

An important decision when designing effective controllers for mobile robots is how much *a priori* knowledge should be imparted to them. Should they attempt to learn all behaviours during the task, or should they begin with a set of pre-engineered actions? Both of these alternatives have considerable drawbacks; starting with no prior knowledge increases task time substantially because the robot has to undergo a learning period during which it is also at risk of damage. However, if it is solely reliant on designer-prescribed behaviours, it has no capacity for learning and adaptation.

The architecture described in this paper takes inspiration from the vertebrate immune system in order to attempt to overcome these problems. The immune system learns to recognize antigens over the lifetime of the individual (Short-Term Learning, STL), but also has knowledge of how to build successful antibodies from gene libraries that have evolved over the lifetime of the species (Long-Term Learning, LTL). This "two timescale" approach can be mimicked by coupling an idiotypic Artificial Immune System (AIS) scheme (STL phase) with a Genetic Algorithm (GA) that rapidly evolves sets of behaviours in simulation (LTL phase) to seed the AIS. This removes any need for hand-designing, permits more scope for creating adaptive solutions, and prevents robots from having to begin a task with no knowledge. The main focus here is describing the idiotypic AIS system (as the GA has already been treated in [1]), and testing whether the seeded system outperforms an unseeded one in both the real and simulated domains. In addition, the role of idiotypic selection in the STL is also examined by trialing systems that do not employ this feature.

The paper is arranged as follows. Section 2 discusses previous idiotypic AIS robot-controllers, and explains the potential benefits of coupling an LTL phase with an idiotypic system. Section 3 describes the test environments and problem used, and Section 4 presents a thorough description of the STL architecture. Section 5 highlights the experimental procedures used and Section 6 reports on and discusses the results obtained. Section 7 concludes the paper.

## 2  Background

The aim of this paper is chiefly to investigate whether there are distinct advantages to integrating LTL strategies (a GA run in fast simulation is used here) with STL strategies. In theory, the LTL phase should be able to provide the STL phase with unbiased (i.e. non-user-designed) starting behaviours, and the STL should permit the continued adaptation of the behaviours as the robot carries out its task in real time.

The STL phase used here is an idiotypic AIS network based on Farmer *et al.*'s [15] model of Jerne's [16] idiotypic network theory. In the model, antibody concentrations are dependent both on the antigens present, and on the other antibodies in the network, i.e. antibodies are suppressed and stimulated by each other as well as being stimulated by antigens. This means that the antibody that best matches the invading antigen is not necessarily selected for execution, which produces a more flexible and dynamic system. The theory has proved popular when designing AIS-based robot control systems, since it potentially allows great variability of robot behaviours (modelled by antibodies) in the face of changing environments (modelled by antigens).

However, past research has mostly been concerned with the structure and evolution of the antibody network, and little attention has been given towards the derivation and design of the antibodies themselves. For example, [3]–[7] all use GAs but evolve only the network links between the antibodies, which are hand-designed, fixed, and small in number. Reference [2] also uses a fixed set of pre-engineered antibodies. In contrast, the LTL phase of this research [1] uses a GA where six basic antibody-types are encoded with a set of six variable attributes that can take many different values, meaning that the system can evolve complete sets of simple but very diverse antibodies. These can then be passed to the STL phase, providing the potential to bestow much greater flexibility to the idiotypic system. In addition, the use of rapid simulations means that the AIS can be seeded within a very realistic time frame (less than twenty minutes) whereas most evolutionary work requires much longer to converge, sometimes even a number of days, which is prohibitive. For example, the systems developed in [8]–[11] have not overcome the unrealistic convergence-time problems.

The most important questions, however, are whether the evolved antibodies can be used effectively in an STL system, and whether such systems can cope with different environments, particularly the real world. Since environments can change, any form of STL needs to be adaptable as well as robust. Previous attempts at fusing STL and LTL include the use of neural networks, for example [12], which proves adaptable to different environments and across different platforms, but the system is trialed using a simple light-switching problem with no obstacles apart from the pen walls. In the experiments described here, more complex problems and much busier environments are employed for testing. In [13] an evolutionary strategy is used for the STL phase. This provides continued adaptation, but deals with a maximum of only 21 behaviour parameters in the LTL phase. Here, behaviours are assembled in a piecewise fashion and from a huge pool of parameters, which should mean greater flexibility. In [14] the two learning phases are implemented simultaneously, but the system is trialed only in simple, structured environments. In addition, the authors claim an evolutionary period of only five minutes, but the results suggest that the robot was unable to avoid the obstacle prior to this. In contrast, the seeded STL system discussed here does not start until it has received the complete sets of GA-derived behaviours, so that it is fully ready to begin the task.

In order to establish that the initial seeding is extremely important in producing a robust and adaptable controller, unseeded systems (i.e. with no LTL phase) that begin with random behaviour sets are also tested. In addition, both the seeded and unseeded systems are run with and without the use of idiotypic effects, to establish the role of the idiotypic mechanism in providing flexibility. A hand-designed controller is also included to investigate how fixed strategies compare with variable ones. It uses a simple random wander to search for the target, a backward turning motion to escape collisions, and it steers the robot in the opposite direction of any detected obstacles. The research thus aims to investigate the following hypotheses:

$H_1$  Seeded STL systems outperform unseeded STL systems.
$H_2$  Seeded STL systems that employ idiotypic effects outperform seeded systems that do not.
$H_3$  Seeded STL systems that employ idiotypic effects outperform fixed, hand-designed strategies.
$H_4$  As long as the LTL-derived behaviours are sufficiently diverse, antibody replacement should not be necessary in the STL phase.

Reference [2] has already provided statistical evidence that idiotypic AIS systems are more effective than similar non-idiotypic ones, but it is restricted to a single robotic platform (Pioneer 3), the simulated domain, and only two different environments. This paper will hence also extend the research in [2] to include a different type of robot (e-puck), more environments, different problems, the real domain, an alternative RL strategy (see section 4.4), and a variable idiotope (see section 4.2).

## 3   Test Environments and Problem

The STL is conducted with an e-puck, a miniature mobile-robot with a small frontal camera and eight infra-red (IR) sensors that can detect the presence of objects up to a distance of about 0.1 m. Both virtual and real environments are used for testing. The simulated environments are worlds that have been designed using Webots [17] software, since the GA employs it, and many modules from the GA can be re-used for the AIS. Webots also permits easy transfer of control from the simulation to the real robot. Two simulated worlds are considered, World 1 (see Fig. 1), and World 2 (see Fig. 2). In these the robot begins south of the central row of pillars and must detect and travel to the blue target-block in the north, avoiding collision with the obstacles, walls and pillars. In addition, a wandering e-puck acts as a dynamic obstacle. Once the robot has arrived at the target, the number of collisions $c$ and the time to complete the task $\tau$ are recorded. The starting positions of the robots and target block are changed automatically after each run.

The real environment consists of a square wooden pen with sides 1.26 m long and 0.165 m high, (see Fig. 3). The mission robot must find and travel to the blue ball located in the pen, avoiding collisions. Once it has found the ball it must stop to signal that the target has been found. The obstacles, robots and ball are randomly placed in different starting positions after each run, so that the environment is slightly different in each case.

The seeded systems all take their starting antibody-sets from those created when the GA is run in the first world described in [1], i.e. a maze-world where the robot must track painted doors in order navigate to the end, (see Fig. 4). This world is employed in the LTL phase to show that the evolved behaviours do not have to be generated using the same environment and problem as in the STL phase.

Webots version 5.7.0 is used, running on GNU/Linux 2.6.9 (CentOS distribution) with a Pentium 4 processor (clock speed 3.6 GHz). For both the real and virtual e-pucks the camera field-of-view is set to 0.3 radians, the pixel width and height to 15 and 3 pixels, and the speed unit for the wheels is set to 0.00683 radians/s.

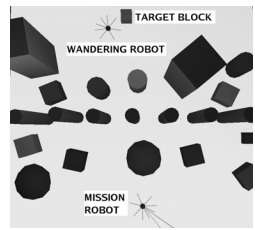

**Fig. 1.** Simulated World 1

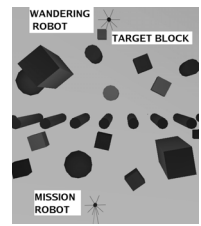

**Fig. 2.** Simulated World 2

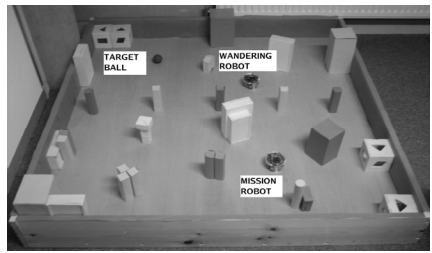

**Fig. 3.** Real World

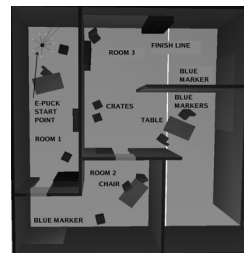

**Fig. 4.** GA Maze World

## 4 System Architecture

### 4.1 Antibodies and Antigens

There are eight antigens indexed 0-7, but only one presents itself at any instant. Either "0 - target unseen" or "1 - target seen" is active when no obstacles are present, (when the maximum IR reading $V_m$ is less than 250). If $V_m$ is between 250 and 2400 then either "2 - obstacle right", "3 - obstacle rear" or "4 - obstacle left" is active. If $V_m$ is 2400 or more then "5 - collision right", "6 - collision rear", or "7 - collision left" presents itself.

There are six basic types of antibody, as listed in Table 1, and each possesses the attributes type $T$, speed $S$, frequency of turn $F$, turn angle $A$, direction of turn $D$, frequency of right turn $R_f$, angle of right turn $R_a$, and cumulative RL-score $L$. However, some types have null values for some attributes, and there are set limits to the values that the attributes can take.

**Table 1.** System Antibody Types

| $T$ | Description | $S$ Speed Units / s | | $F$ % of time | | $A$ % reduction in speed of one wheel | | $D$ Either left or right | | $R_f$ % of time | | $R_a$ % reduction in right wheel-speed | |
|---|---|---|---|---|---|---|---|---|---|---|---|---|---|
| | | MIN | MAX | MIN | MAX | MIN | MAX | 1 | 2 | MIN | MAX | MIN | MAX |
| 0 | Wander single | 50 | 400 | 10 | 90 | 10 | 110 | L | R | - | - | - | - |
| 1 | Wander both | 50 | 400 | 10 | 90 | 10 | 110 | - | - | 10 | 90 | 10 | 110 |
| 2 | Forward turn | 50 | 400 | - | - | 20 | 200 | L | R | - | - | - | - |
| 3 | Static turn | 50 | 100 | - | - | 100 | 100 | L | R | - | - | - | - |
| 4 | Reverse turn | 300 | 400 | - | - | 20 | 200 | L | R | - | - | - | - |
| 5 | Track markers | 50 | 400 | - | - | 0 | 30 | - | - | - | - | - | - |

### 4.2 Creating the Paratope and Idiotope Matrices

An antibody set consists of eight behaviours, one for each antigen, and five distinct antibody sets are used. The 40 antibodies in the system can hence be represented as $A_{ij}$, $i = 0, \ldots, x\text{-}1, j = 0, \ldots, y\text{-}1$, where $x$ is the number of sets and $y$ is the number of antigens. For the seeded systems the evolved sets of antibody attribute values, their associated task completion times $\tau_i$, and numbers of collisions $c_i$ are read in from the file previously created when the GA was run. The STL system then calculates the relative fitness of each antibody set $\mu_i$ from:

$$\mu_i = \frac{1}{(\tau_i + \rho c_i) \sum_{k=0}^{x-1} (\tau_k + 8c_k)^{-1}}, \tag{1}$$

where $\rho = 8$ to give $c$ equal weight compared to $\tau$. It then produces a matrix of RL scores $P_{ij}$, which are analogous to antibody paratope values, as the scores represent a comparative estimate of how well each antibody matches its antigen, see [2]. The elements of $P_{ij}$ are calculated by multiplying the antibody's final RL score $L_{ij}$ by the relative fitness $\mu_i$ of its set, and scaling approximately to between 0.00 and 1.00 using:

$$P_{ij} = \frac{L_{ij}\mu_i}{\varphi}. \tag{2}$$

Taking $\varphi = 20$ works here since the approximate maximum value $L_{ij}\mu_i$ can take is 20. For the unseeded systems the five antibody sets are generated at the start of the STL phase, by randomly choosing behaviour types and their attribute values. The initial elements of $P_{ij}$ are also randomly generated, but always lie between 0.25 and 0.75 to try to limit any initial biasing of the selection.

For both seeded and unseeded systems, a matrix $I_{ij}$ (analogous to a matrix of idiotope values, see [2]) is created by comparing the individual paratope matrix elements $P_{ij}$ with the mean element value for each of the antigens $\sigma_j$. This is given by:

$$\sigma_j = \frac{\sum_{i=0}^{x-1} P_{ij}}{x}. \tag{3}$$

If $P_{ij}$, $i = 0, \ldots, x\text{-}1$ is less than $\sigma_j$, then an idiotope value $I_{ij}$ of 1.0 is assigned, otherwise a value of zero is given. However, only one antibody in each set may have a non-zero idiotope. If more than one has a non-zero value, then one of them is selected at random and all the others are set back to zero. This avoids over-stimulation or over-suppression of antibodies.

The paratope matrix is adjusted after every iteration; first, because the active antibody's paratope value either increases or decreases, depending on the RL score awarded, and second, because all the paratope values are then re-calculated, so that the $\sigma_j$ values are changed back to the initial mean values. The adjustment is given by:

$$P_{ij_{t+1}} = P_{ij_t} \frac{\sigma_{j_0}}{\sigma_{j_t}}, \tag{4}$$

where $\sigma_{j_0}$ represents the initial means and $\sigma_{j_t}$ represents the temporary means obtained after the active antibody has been scored. This adjustment helps to eliminate the problems that occur when useful antibodies end up with zero $P_{ij}$ values. The idiotope is re-calculated, based on the latest $P_{ij}$ values, after every 120 sensor readings, i.e. every 3.84 s, since the sensors are read every 32 milliseconds.

### 4.3 Antibody Selection Process

At the start of the STL phase each antibody has 1000 clones in the system, but the numbers fluctuate according to a variation of Farmer's equation:

$$N_{im_{(t+1)}} = bS_{im_{(t)}} + N_{im_{(t)}}(1 - k_3), \tag{5}$$

where $N_{im}$ represents the number of clones of each antibody matching the invading antigen $m$. $S_{im}$ is the current strength-of-match of each of these antibodies to $m$, $b$ is a scaling constant and $k_3$ is the death rate constant, (see [2] for further details). The concentration $C_{ij}$ of every antibody in the system consequently changes according to:

$$C_{ij} = \frac{\Phi N_{ij}}{\sum_{k=0}^{x-1}\sum_{l=0}^{y-1} N_{kl}}, \qquad (6)$$

where $\Phi$ is another scaling factor that can be used to control the levels of inter-antibody stimulation and suppression (25 is used here).

The antibody selection process comprises three stages for idiotypic selection, but only one stage if idiotypic selection is not used. First, the sensors are read to determine the index of the presenting antigen $m$, and an appropriate antibody is selected from those available for that antigen. More specifically, the system chooses from antibodies $A_{im}$, $i = 0, \ldots, 4$, by examining the paratope values $P_{im}$. The antibody $\alpha$ with the highest of these paratope values is chosen as the first stage winner. If the index of the winning antibody set is denoted as $n$, then $\alpha = A_{nm}$. If idiotypic effects are not considered $\alpha$ carries out its action, and is assessed by RL, see section 4.4.

If an idiotypic system is used, then the stimulatory and suppressive effects of $\alpha$ on all the antibodies in the repertoire must be considered. As detailed in [2], this involves comparing the idiotope of $\alpha$ with the paratopes of the other antibodies to determine how much each is stimulated, and comparing the paratope of $\alpha$ with the idiotopes of the others to calculate how much each should be suppressed. Here, idiotypic selection is governed by equations (7)-(10), which are based on those in [2]. Equation (7) concerns the increase in strength-of-match value $\varepsilon_{im}$ when stimulation occurs,

$$\varepsilon_{im} = k_1 \sum_{j=0}^{y-1}(1 - P_{ij}) I_{nj} C_{ij} C_{nj}, \qquad (7)$$

where $k_1$ is a constant that determines the magnitude of any stimulatory effects. The formula for the reduction in strength-of-match value $\delta_{im}$ when suppression occurs is:

$$\delta_{im} = k_2 \sum_{j=0}^{y-1} P_{nj} I_{ij} C_{ij} C_{nj}, \qquad (8)$$

where $k_2$ governs the suppression magnitude. Hence, the strength-of-match after the second selection-stage $(S_{im})_2$ is given by:

$$(S_{im})_2 = (S_{im})_1 + \varepsilon_{im} - \delta_{im}, \qquad (9)$$

where the initial strength-of-match $(S_{im})_1$ for each antibody is taken as the current $P_{im}$ value. After the $(S_{im})_2$ values are calculated, the numbers of clones $N_{im}$ are adjusted using (1) and all concentrations $C_{ij}$ are re-evaluated using (2). The third stage calculates the activation $\lambda$ of each antibody in the sub-set $A_{im}$ from

$$\lambda_{im} = C_{im} (S_{im})_2. \tag{10}$$

The third-stage winning antibody $\beta$ has the highest $\lambda$ value. If $p$ is the index of $\beta$'s antibody set, then $\beta = A_{pm}$. When idiotypic selection is used, $\beta$ carries out its action and it is $\beta$ that is scored using RL rather than $\alpha$, although $\alpha$ and $\beta$ will be the same if $n = p$.

**4.4 Reinforcement Learning and Antibody Replacement**

Reinforcement learning scores the performance of an antibody by comparing old and new environmental information. Here, the antibody used in the previous iteration $A_{t-1}$ is assessed by examining the current and previous antigen codes $m_t$ and $m_{t-1}$. Table 2 shows the RL score $r$ awarded for each possible combination. The final score given is dependent on how many environmental changes have taken place, and whether the change is negative or positive, for example, moving away from an obstacle is a valuable improvement, and would yield a positive component of 0.1. The maximum cumulative-RL-score (or $P_{ij}$ value) allowed is 1.00, and the minimum $P_{ij}$ value is 0.00.

The $P_{ij}$ values are also affected when the antigen code has remained at 0 for more than 250 iterations, as this means that the robot is spending too much time wandering and has not found anything. It is important to recognize this behaviour as negative, as otherwise robots may be circling around on the spot, never achieving anything, but receiving constant rewards. The non-idiotypic case reduces the cumulative-RL-score by 1.0, and the idiotypic case reduces it by 0.5, as pre-trials have shown that non-idiotypic robots require a more drastic change to break out of repeated behaviour cycles. The same $P_{ij}$ adjustments are also made if there have been more than 15 consecutive obstacle encounters, as this may indicate that a robot is trapped.

Following RL, the paratope values are scaled using (4). In the case of the unseeded trials, replacement occurs for all antibodies with $P_{ij}$ less than 0.1. When this takes place, a new antibody is created by randomly choosing a behaviour type and its attribute values. Antibody replacement is not used in the seeded systems, since $H_4$ is directly concerned with establishing whether this is necessary.

**Table 2.** Reinforcement Scores

| Antigen code | | $r$ score | Reinforcement status |
|---|---|---|---|
| Old | New | | |
| 0 | 0 | 0.05 | Reward – No obstacles encountered |
| 1 | 0 | -0.10 | Penalize - Lost sight of marker |
| 2-7 | 0 | 0.10 | Reward - Avoided obstacle |
| 0 | 1 | 0.10 | Reward - Found marker |
| 1 | 1 | 0.00 to 0.05 | Reward – Kept sight of marker (Score depends on orientation of marker with respect to robot) |
| 2-7 | 1 | 0.20 | Reward - Avoided obstacle and gained or kept sight of marker |
| 0 | 2-7 | -0.05 | Penalize – Encountered obstacle |
| 1 | 2-7 | -0.05 | Penalize – Encountered obstacle |
| 2-7 | 2-7 | -0.40 to 0.50 | Reward or Penalize (Score depends on several factors) |

## 5   Experimental Procedures

Before any of the seeded STL-phase tests take place, the GA is run once in the maze world, in accordance with the procedures described in [1], to obtain the initial seeding. Five independent populations of ten robots and a mutation rate of 5% are used, as recommended in [1]. Following this, 30 STL trials are performed in each of the two simulated worlds, and 20 are completed in the real world. This is done for each of the following systems; seeded with idiotypic effects, seeded with RL only, unseeded with idiotypic effects, unseeded with RL only, and a hand-designed controller. In the unseeded simulated-worlds two separate sets of experiments are conducted with two different initially-random behaviour sets $R_1$ and $R_2$. The real-world unseeded experiments use only $R_1$ since they have to run in real time and are hence much more time consuming to carry out.

In the idiotypic systems $b$ is set to 100, $k_3$ is set to zero, and $k_1$ and $k_2$ are set at 0.85 and 1.10 respectively. These values are chosen in order to yield a mean idiotypic difference rate of approximately 20%, as this is advised in [2]. N. B. An idiotypic difference occurs when the antibodies $\alpha$ and $\beta$ are different. For all experiments, the time taken $\tau$ and the number of collisions $c$ are capped at 4000 s and 100 respectively. Any runs that exceed either of these limits are counted as failed runs. The fitness $f$ is calculated as:

$$f = \frac{\rho c + \tau}{2}, \qquad (11)$$

where $\rho = 8$ as before. A run finishes when the robot has detected three consecutive instances of more than 40 blue pixels in the ball image, so that it is "aware" of having found its target. Standard two-tailed $t$-tests are applied to compare the various systems, and differences are accepted as significant at the 99% level only.

## 6   Results

Table 3 shows the mean $c$, $\tau$, and $f$ values for each of the systems in each of the worlds, and Table 4 presents the significant difference levels when the systems are compared. Table 5 displays the failure rates, indicating the percentage of failures due to an excessive number of collisions, running out of time, and overall.

In all of the worlds, both simulated and real, the system with the lowest $c$, fastest $\tau$, and best $f$ is the seeded idiotypic system. When compared with the unseeded systems it is significantly better in all cases, i.e. for all of the metrics, in all the worlds, and irrespective of whether the unseeded systems use idiotypic effects, or which random behaviour set is used.

However, when the non-idiotypic seeded system is compared with the unseeded systems, although its performance is better in all cases, it is not always significantly better. Most of the significant differences arise when comparing seeded and unseeded systems that do not use idiotypic effects. In these cases, $c$ is always significantly better for the seeded system, and, when $R_2$ is used in unseeded system, the seeded one is always significantly better. When the unseeded system employs idiotypic effects and the seeded system does not, there is a marked drop in the percentage of significant differences, although many of the collision comparisons are significantly better for the seeded system.

When the seeded idiotypic system is compared with the seeded non-idiotypic system, the idiotypic system performs better in all cases, and significantly better in most cases. However, when the unseeded systems are compared in this way, although the idiotypic system consistently performs better, none of the differences are significant.

The seeded idiotypic system surpasses the hand-designed controller in all cases (except for a tie in $c$ in Simulated World 2), and more than half of these differences are significant overall. However, in the real world all of the differences are significant. It appears that the hand-designed controller performs very well in the simulator in terms of $c$, but poorly for $\tau$, whereas in the real world it performs badly for both of these metrics. The seeded idiotypic system works well in the real world and in the simulator for both $c$ and $\tau$. In fact, in the real world it proves significantly better than all of the other systems trialed, for all metrics.

**Table 3.** Mean $c$, $\tau$, and $f$. (S = seeded, U = unseeded, IE = idiotypic effects, RL = reinforcement learning, HDC = hand-designed controller)

| System | Set | Simulated World 1 | | | Simulated World 2 | | | Real World | | |
|---|---|---|---|---|---|---|---|---|---|---|
| | | $c$ | $\tau$ | $f$ | $c$ | $\tau$ | $f$ | $c$ | $\tau$ | $f$ |
| SIE | - | **1** | **562** | **284** | **2** | **659** | **336** | **5** | **283** | **161** |
| SRL | - | 8 | 1298 | 679 | 4 | 1113 | 573 | 23 | 904 | 544 |
| UIE | $R_1$ | 26 | 1513 | 862 | 26 | 1530 | 868 | 96 | 1384 | 1074 |
| URL | $R_1$ | 45 | 2150 | 1253 | 35 | 1732 | 1006 | 100 | 1678 | 1239 |
| UIE | $R_2$ | 20 | 1720 | 941 | 48 | 1578 | 981 | - | - | - |
| URL | $R_2$ | 35 | 2214 | 1246 | 54 | 2137 | 1285 | - | - | - |
| HDC | - | 2 | 1362 | 688 | **2** | 1256 | 636 | 44 | 1439 | 897 |

**Table 4.** Significance Levels (S = seeded, U = unseeded, IE = idiotypic effects, RL = reinforcement learning, HDC = hand-designed controller)

| Systems | | Set | Simulated World 1 | | | Simulated World 2 | | | Real World | | |
|---|---|---|---|---|---|---|---|---|---|---|---|
| | | | $c$ | $\tau$ | $f$ | $c$ | $\tau$ | $f$ | $c$ | $\tau$ | $f$ |
| SIE | SRL | - | 100 | 100 | 100 | 98 | 96 | 97 | 99 | 99 | 100 |
| SIE | HDC | - | 85 | 100 | 100 | 33 | 97 | 97 | 100 | 100 | 100 |
| SIE | UIE | $R_1$ | 100 | 100 | 100 | 100 | 100 | 100 | 100 | 100 | 100 |
| SIE | URL | $R_1$ | 100 | 100 | 100 | 100 | 100 | 100 | 100 | 100 | 100 |
| SIE | UIE | $R_2$ | 99 | 100 | 100 | 100 | 100 | 100 | - | - | - |
| SIE | URL | $R_2$ | 100 | 100 | 100 | 100 | 100 | 100 | - | - | - |
| SRL | UIE | $R_1$ | 98 | 49 | 72 | **99** | 83 | 92 | **100** | 85 | **99** |
| SRL | URL | $R_1$ | 100 | 99 | 100 | 100 | 94 | **98** | **100** | 96 | **100** |
| SRL | UIE | $R_2$ | 91 | 82 | 89 | **100** | 86 | **98** | - | - | - |
| SRL | URL | $R_2$ | **100** | **99** | **100** | 100 | **100** | **100** | - | - | - |
| UIE | URL | $R_1$ | 87 | 90 | 93 | 59 | 44 | 52 | 68 | 53 | 57 |
| UIE | URL | $R_2$ | 82 | 81 | 87 | 40 | 86 | 84 | - | - | - |

**Table 5.** Percentage Failure Rates (S = seeded, U = unseeded, IE = idiotypic effects, RL = reinforcement learning, HDC = hand-designed controller)

| System | Set | Simulated World 1 (%) | | | Simulated World 2 (%) | | | Real World (%) | | | Mean (%) | | |
|---|---|---|---|---|---|---|---|---|---|---|---|---|---|
| | | $c$ | $\tau$ | Tot | $c$ | $\tau$ | Tot | $c$ | $\tau$ | Tot | $c$ | $\tau$ | Tot |
| SIE | - | **0** | **0** | **0** | **0** | **0** | **0** | **0** | **0** | **0** | **0** | **0** | **0** |
| SRL | - | **0** | 3 | 3 | **0** | 7 | 7 | 10 | 5 | 10 | 3 | 5 | 7 |
| UIE | $R_1$ | 23 | 17 | 30 | 20 | 13 | 23 | 95 | 10 | 95 | 46 | 13 | 49 |
| URL | $R_1$ | 43 | 30 | 57 | 33 | 23 | 47 | 100 | 20 | 100 | 59 | 24 | 68 |
| UIE | $R_2$ | 17 | 20 | 37 | 43 | 17 | 43 | - | - | - | 30 | 18 | 40 |
| URL | $R_2$ | 30 | 30 | 47 | 50 | 27 | 53 | - | - | - | 40 | 28 | 50 |
| HDC | - | **0** | 20 | 20 | **0** | 17 | 17 | 10 | 25 | 35 | 3 | 21 | 24 |

Furthermore, the seeded idiotypic system is the only scheme that consistently displays a 0% failure rate. Failure rates are reasonably low (7% overall) for the non-idiotypic seeded system, but reach unacceptable proportions for the hand-designed controller (24% overall) and the idiotypic unseeded system (49% and 40% overall). The non-idiotypic unseeded system is clearly the worst option with overall fail rates of 68% and 50%. Moreover, the actual number of collisions for failing robots is often of the order of thousands for unseeded real-world systems, which renders the method entirely unsuitable.

These observations represent very strong statistical evidence in support of $H_1$ and $H_3$, i.e. they recommend the use of GA-seeded systems over both unseeded systems and fixed, user-designed systems. In particular, there is over-whelming statistical evidence in favour of using a seeded idiotypic system over any unseeded system, with all tests proving highly significant. In addition, the results provide some evidence to uphold $H_2$, since robot performance appears to be further enhanced by incorporating an idiotypic network into the STL architecture. In the seeded idiotypic system the GA provides immediate knowledge of how to begin the task, and the idiotypic AIS permits it to change and adapt its behaviour as the need arises. Without idiotypic effects, the seeded system has the same initial knowledge, but relies only on RL for adaptation, so it is less flexible. Although the hand-designed controller has built-in initial knowledge, it also proves inferior because of its inability to change the way it responds to an antigen. In contrast, the unseeded systems have no initial knowledge, and must acquire their abilities during the STL phase. This is a very slow process, even when idiotypic selection is used, because the search space is probably much too large given the time frame for completing the task. Moreover, the mechanism by which antibodies are replaced is not well developed; the robot is forced to select a random behaviour when it rejects an antibody, and could hence still be using random antibodies during the latter stages of task completion.

The results also demonstrate that behaviours derived in GA simulations can transfer extremely well to the real world, even when the simulated and real environments are very different. In addition, the tests show that the superiority of idiotypic AIS systems over RL-only systems (suggested in [2]) can be extended to the real world, other simulated worlds, and a different robotic platform. These experiments also uphold $H_4$, since the seeded idiotypic system exhibits a 0% fail rate in all cases, suggesting that antibody replacement is not necessary when adequate seeding and a sufficiently adaptive strategy are in place.

## 7 Conclusions

This paper has described merging LTL (an accelerated GA run in simulation), with STL (an idiotypic AIS scheme), in order to seed the AIS with sets of very diverse behaviours that can work together to solve a mobile-robot target-finding problem. Results have shown that such seeded systems consistently perform significantly better than unseeded systems, and have also provided strong statistical evidence that the idiotypic selection process contributes towards this improved performance. The fusion of the two learning timescales has been shown to provide a rapid and realistic method for training robots in simulation, and an adaptable and robust system for carrying out real world activities.